# Statistical power for cluster analysis

Edwin S. Dalmaijer, Camilla L. Nord, & Duncan E. Astle

*MRC Cognition and Brain Sciences Unit, University of Cambridge, Cambridge, United Kingdom*

**Corresponding author**

Dr Edwin S. Dalmaijer, MRC Cognition and Brain Sciences Unit, 15 Chaucer Road, Cambridge, CB2 7EF, United Kingdom. Telephone: 0044 1223 769 447. Email: <u>edwin.dalmaijer@mrc-cbu.cam.ac.uk</u>

# Abstract

**Background.** Cluster algorithms are gaining in popularity in biomedical research due to their compelling ability to identify discrete subgroups in data, and their increasing accessibility in mainstream software. While guidelines exist for algorithm selection and outcome evaluation, there are no firmly established ways of computing a priori statistical power for cluster analysis. Here, we estimated power and classification accuracy for common analysis pipelines through simulation. We systematically varied subgroup size, number, separation (effect size), and covariance structure. We then subjected generated datasets to dimensionality reduction approaches (none, multi-dimensional scaling, or uniform manifold approximation and projection) and cluster algorithms (k-means, agglomerative hierarchical clustering with Ward or average linkage and Euclidean or cosine distance, HDBSCAN). Finally, we directly compared the statistical power of discrete (k-means), "fuzzy" (c-means), and finite mixture modelling approaches (which include latent class analysis and latent profile analysis).

**Results.** We found that clustering outcomes were driven by large effect sizes or the accumulation of many smaller effects across features, and were mostly unaffected by differences in covariance structure. Sufficient statistical power was achieved with relatively small samples (N=20 per subgroup), provided cluster separation is large ($\Delta=4$). Finally, we demonstrated that fuzzy clustering can provide a more parsimonious and powerful alternative for identifying separable multivariate normal distributions, particularly those with slightly lower centroid separation ($\Delta=3$).

**Conclusions.** Traditional intuitions about statistical power only partially apply to cluster analysis: increasing the number of participants above a sufficient sample size did not improve power, but effect size was crucial. Notably, for the popular dimensionality reduction and clustering algorithms tested here, power was only satisfactory for relatively large effect sizes (clear separation between subgroups). Fuzzy clustering provided higher power in multivariate normal distributions. Overall, we recommend that researchers 1) only apply cluster analysis when large subgroup separation is expected, 2) aim for sample sizes of N=20 to N=30 per expected subgroup, 3) use multi-dimensional scaling to improve cluster separation, and 4) use fuzzy clustering or mixture modelling approaches that are more powerful and more parsimonious with partially overlapping multivariate normal distributions.

# Background

Cluster analyses are unsupervised machine-learning algorithms that aim to delineate subgroups in datasets, characterised by discrete differences. For example, while opinions on food tend to vary on a continuous scale from approval to disapproval, certain foods (yeast-based spread Marmite or salted liquorice) anecdotally inspire a much more discrete distribution: people either love them or they hate them, and no lukewarm opinions exist. In medicine, such discrete differences can be of high clinical relevance, e.g. in the diagnosis of polycystic ovary syndrome, where serum testosterone occurs in two discrete (but partially overlapping) subgroups: relatively high for patients, and relatively low for non-patients (1). Traditionally, such groupings are identified on the basis of known differences, like a diagnosis. However, clustering algorithms provide a method for the data-driven identification of subgroups without the need for a-priori labelling.

Due to this unique property, cluster analysis has gradually increased in popularity compared to traditional statistical analyses (Figure 1). This was likely helped by the increase in desktop computing power, and the implementation of cluster algorithms in popular statistics software like Python (2), R (3), JASP, Matlab, Stata, and SPSS. One recent example can be found in type-2 diabetes, where researchers have aimed to use data-driven clustering approaches to delineate subgroups that could have different disease progression and risk profiles, and potentially benefit from different types of treatment. Ongoing discussion centres on whether four (4–6) or five (7) groups exist, and what characterises each cluster. Other examples include the identification of protein communities involved in cancer metastasis (8), responder types to cancer treatment (9), Parkinson's disease subtypes (10), brain types (11), and behavioural phenotypes (12–16).

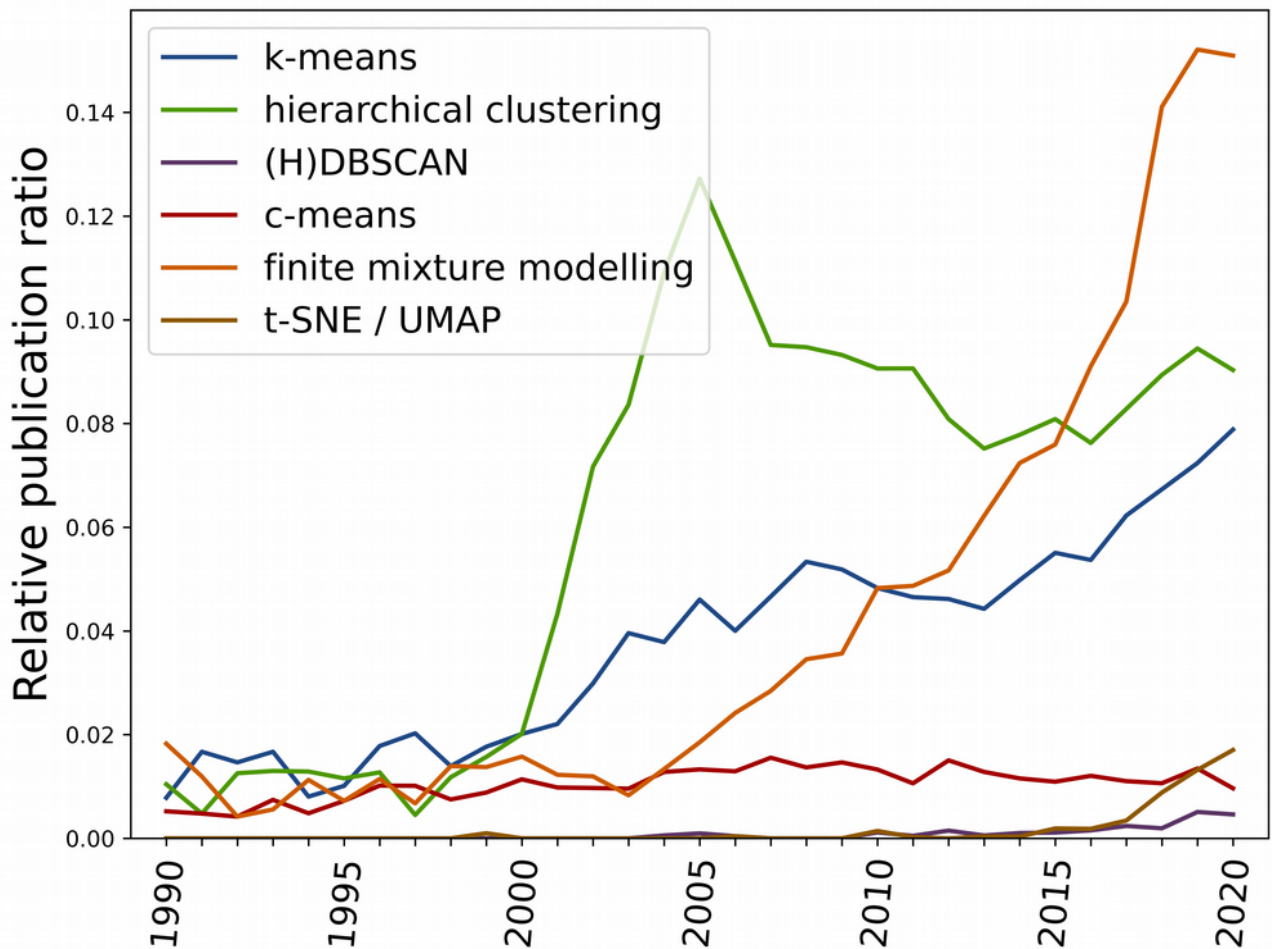

*Figure 1.* *The increase in publications indexed by PubMed that mention a keyword specific to cluster analyses relative to the number of publications that mention a traditional statistical test. Particularly sharp increases can be seen for finite mixture modelling (which includes latent class and latent profile analysis) and k-means. Illustration generated using Bibliobanana (17). See Methods for details on exact search terms.*

A major concern with the adoption of clustering techniques in medical research, is that they are not necessarily employed on the type of data for which they were developed or validated. In typical biomedical data, even well-separated groups (like the aforementioned polycystic ovary syndrome patients and controls) usually show partial overlap. Observations from each group form a multivariate normal distribution (a Gaussian distribution over more than one feature), and together they make up a multimodal distribution (a distribution with more than one peak). Such multimodal distributions are mostly ignored in clustering tutorials that provide idealised data from strongly separated high-dimensional "blobs" (Figure 2, left column), or from features measured in clearly distinct groups such as different species of flowers (18) (Figure 2, middle column). Even studies aimed at identifying factors that impact cluster algorithm performance have used, perhaps unrealistically, large cluster separation (19,20). In fact, the popular "gap statistic" evaluation metric is explicitly designed for well-separated and non-overlapping clusters (21). In sum, real biomedical data frequently takes the form of multivariate normal distributions that are not particularly well separated (Figure 2, right column), and it is unclear what this means for the suitability of cluster analyses.

One potential way around this problem is to simply argue that when subgroups are not as well-separated, "the notion of a cluster is not any more well defined in the literature" (21). This sidesteps the issue, and ignores the ongoing philosophical debate on what constitutes a cluster (22) (not to mention the fact that overlap can be introduced or exaggerated by measurement error). Furthermore, some of the cited examples from the biomedical literature suggest that researchers have already decided that partially overlapping subgroups are valid targets for cluster analysis. Hence, the best pragmatic approach would be to investigate when cluster analysis can or cannot help to identify subgroups in a dataset.

Several excellent resources exist on the various steps involved in typical cluster analyses, and the benefits and potential pitfalls of algorithms and metrics (23–25). Our main aim was not to provide another comprehensive overview and comparison of all techniques, but rather to provide the necessary pragmatic information for researchers who are planning to incorporate cluster analysis in their study. Specifically, we investigated the statistical power and accuracy of several of the most popular approaches; and offer suggestions on when cluster analysis is applicable, how many observations are required, and which types of techniques provide the best statistical power.

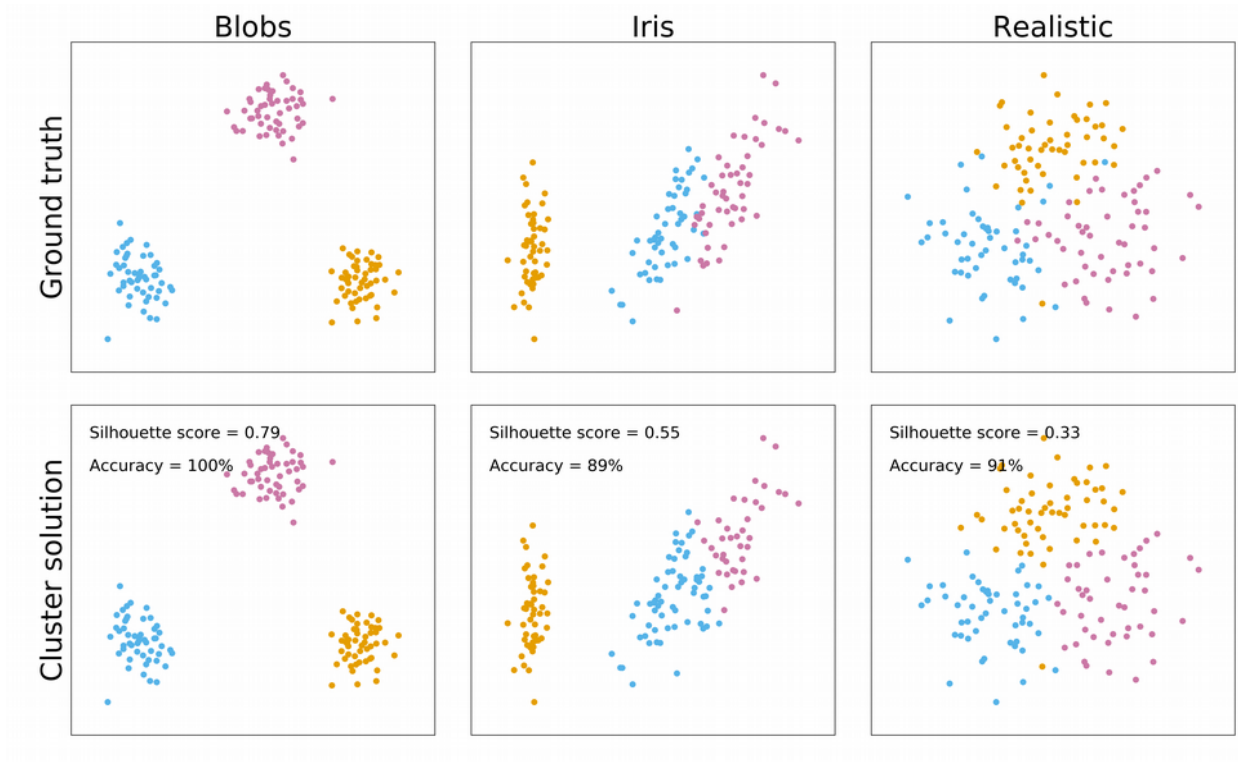

*Figure 2* – *This figure shows three datasets in the top row, each made up out of 150 observations that fall in three equally sized clusters. While the datasets are made up of 4 features, the plotted data is a two-dimensional projection through multi-dimensional scaling (MDS). The left column presents simulated "blobs" as they are commonly used in clustering tutorials, the middle column presents the popular Iris dataset, and the right presents more realistic multivariate normal distributions. The bottom row presents the outcome of k-means clustering, showing good classification accuracy for all datasets, but only reliable cluster detection (silhouette coefficient of 0.5 or over) for blobs and the Iris dataset, but not the more realistic scenario.*

## A typical cluster analysis

A typical cluster analysis pipeline consists of three different steps: dimensionality reduction, cluster identification, and outcome evaluation. Datasets typically consist of many observations (e.g.

participants or cells) that are characterised by many features (e.g. age, height, weight, etc.). Each feature can be thought of as an additional dimension to the data, making it hard to visualise all data comprehensively in a two-dimensional plot. In addition, clustering algorithms suffer from the "curse of dimensionality", the tendency to perform worse when many features are included (26). Both of these issues can be addressed with dimensionality reduction algorithms that project many features into a low-dimensional space.

The next step is the identification of subgroups, for which many techniques exist. Common across all techniques is to find clusters, and allocate individual observations to those clusters. This can be done on the basis of distance to a cluster centroid (e.g. in k-means), dissimilarity and linkage between groups of observations (e.g. agglomerative hierarchical clustering), or density (e.g. HDBSCAN). Not all techniques assign all observations to a cluster. In addition, there are "fuzzy" techniques (e.g. c-means) that determine for each observation how much it aligns with each cluster. Finally, there are (Gaussian) finite mixture modelling approaches that aim to identify the underlying model that produced a dataset, with each "cluster" being a distribution with a centre and spread. This includes latent profile analysis (for continuous data) and latent class analysis (for binary data). In essence, mixture models quantify the probability of an observation belonging to a cluster, which means the outcome is conceptually similar to fuzzy clustering.

The final step is to quantify the outcome of the cluster analysis. A typical approach is to first determine whether a dataset shows reliable clustering, and then to identify the best-fitting number of clusters. There are many different techniques and metrics for dimensionality reduction, clustering, and outcome evaluation; and more keep being developed. Our aim here is not to assess the latest algorithms, but to investigate the most popular approaches (Figure 1). These are described in further detail in the Methods section.

# Statistical power in cluster analysis

Statistical power is the probability that a test can correctly reject the null hypothesis if the alternative hypothesis is true. In other words, it is the probability of finding a true positive. In many cases, researchers test for differences between groups, or relationships between variables. The first aim is to establish the presence or absence of an effect (e.g. a difference of a relationship); this is a binary decision. If there is indeed an effect, the second aim is to establish its magnitude. Power describes the probability of correctly identifying the presence of an effect. Power increases as a function of this magnitude, and typically also as a function of sample size. Simply put: larger effects are easier to find, and more samples generally make it more likely to find an effect.

Analogously, in cluster analysis the aim is to establish whether a dataset contains subgroups, or comprises a single group. We thus define statistical power in cluster analysis as the probability of correctly that subgroups are present. If this is the case, the next aim is to establish how many clusters are present within the data, and to what extent the cluster membership of individual observations can be accurately classified. An open question is whether traditional intuitions about statistical power hold up in cluster analysis, specifically whether power increases as a function of both effect size and sample size.

It has been suggested that the ubiquity of low statistical power means that "most published research findings are false" (27)., because it reduces our ability to distinguish true and false positives. As an illustration: the majority of neuroscience studies have a very low power with a median 21% (i.e. only 21 out of 100 true effects would be detected) (28), although it has been suggested this varies notably by sub-discipline (29). To help prevent the proliferation of under-powered studies in the era of modern statistics ("data science"), it is important that the statistical power of tools like cluster analysis is investigated.

## The current study

Here, we explore what factors affect cluster algorithms' ability to delineate subgroups by simulating datasets. Three separate simulation analyses were run (Table 1). Crucially, the methods that we tested in these analyses are not the most recent, but rather the most popular methods among the most well-used approaches in the biomedical sciences (see Figure 1).

The first analysis aimed to identify factors that contribute to the separation of cluster centroids. We simulated a range of datasets that varied in the number and relative size of subgroups, covariance structures for each subgroup, and the overall separation between clusters (we independently varied the number of features that were different between subgroups, and the effect size of each difference). We subjected simulated datasets to analysis pipelines that varied in dimensionality reduction and clustering algorithms. Specifically, we used no dimensionality reduction, multi-dimensional scaling (MDS) (30), or uniform manifold approximation and projection (UMAP) (31); and after this we subjected the original or reduced data to k-means (32), agglomerative (hierarchical) clustering (33), or HDBSCAN (34).

Because it became apparent that some factors did not contribute to cluster analysis outcomes, we varied fewer factors for later analyses. In short, covariances between subgroups did not impact outcomes, and agglomerative clustering produced highly similar results to k-means. Furthermore, dimensionality reduction non-linearly (but predictably) scaled the difference between cluster centroids, and it was this difference that ultimately drove outcomes. Hence, we opted for varying subgroup separation, and employed a more narrow set of cluster algorithms.

In the second simulation analysis, we investigated statistical power. To this end, we varied sample size, number and relative size of subgroups, effect size (cluster centroid separation), and cluster algorithm (k-means, HDBSCAN, c-means).

Finally, we note that discrete clustering methods like k-means assign observations to only one cluster. However, particularly in partially overlapping multivariate normal distributions, this leads to overconfident assignment of observations that lie roughly between two centroids. A solution implemented in some algorithms, notably HDBSCAN, is to leave some observations unassigned. However, this does not do justice to the real situation, which is that we can have some confidence that observations halfway between two centroids can be assigned to either cluster, but not to any other cluster with a more distant centroid. Fuzzy clustering allows for exactly this kind of proportionally shared assignment, and hence is perhaps more parsimonious with real data. To explore whether this translates into better statistical power, we compare k-means (discrete), c-means (fuzzy), and Gaussian finite mixture modelling (outcomes conceptually similar to fuzzy).

**Table 1**

*Summary of the simulation analyses that were conducted, and the variables that were varied in each set of simulations. Each unique combination of listed features was simulated. "Ward" and "cosine" refer to agglomerative (hierarchical) clustering, using Ward linkage and Euclidean distance or average linkage and cosine distance, respectively. "Mixture model" refers to finite Gaussian mixture modelling.*

| Analysis | N | k | Effect size | Covariance | Dimensionality reduction | Cluster algorithms |
|---|---|---|---|---|---|---|
| **1) What drives cluster separation** | 1000 | - 2 (10/90%)<br>- 2 (equal)<br>- 3 (equal) | Δ=0.3 – 8.1 | *15 features*<br>- None<br>- Random<br>- Mixed | - None<br>- MDS<br>- UMAP | - K-means<br>- Ward<br>- Cosine<br>- HDBSCAN |
| **2) Statistical power** | 10 – 160 | - 2 (10/90%)<br>- 2 (equal)<br>- 3 (equal)<br>- 4 (equal) | Δ=1 – 10 | *2 features*<br>- None | - None | - K-means<br>- HDBSCAN<br>- C-means |
| **3) Discrete versus fuzzy clustering** | 120 | - 1<br>- 2 (equal)<br>- 3 (equal)<br>- 4 (equal) | Δ=1 – 10 | *2 features*<br>- None | - None | - K-means<br>- C-means<br>- Mixture model |

# Results

## Cluster centroid separation

Using the ground truth cluster membership, we computed the distance Δ between simulated datasets' cluster centroids (after projection into two dimensions) as the Euclidean distance between the average positions of observations within each cluster in reduced space. For three-cluster datasets, we computed the distance between the "middle" cluster and another cluster's centroid to obtain the smallest between-cluster distance.

Centroid distance Δ is plotted as a function of both within-feature effect size δ and the proportion of different features after dimensionality reduction through MDS (Figure 3) and UMAP (Figure 4). The visualised datasets also differed in number of clusters (two unequally sized, two equally sized, or three equally sized), and covariance structure (no covariance, random covariance, or different covariances between clusters), adding up to 180 datasets per figure.

Differences between the number and size of clusters or the covariance structures is negligible for MDS. Overall, cluster separation increases as a function of higher differences between clusters within each feature, and the proportion of different features. The same is true for UMAP, although its outcomes are non-linear and more variable.

In sum, MDS dimensionality reduction shows a steady increase in cluster separation with increasing within-feature effect size and proportion of different features, whereas UMAP shows improved separation only at large differences (Cohen's $d = 2.1$ within each feature) or at large proportions of different features. Crucially, clusters with different covariance structures (3-factor and 4 factor; of 3-factor, 4-factor, and random) but similar mean vectors do not show clear separation.

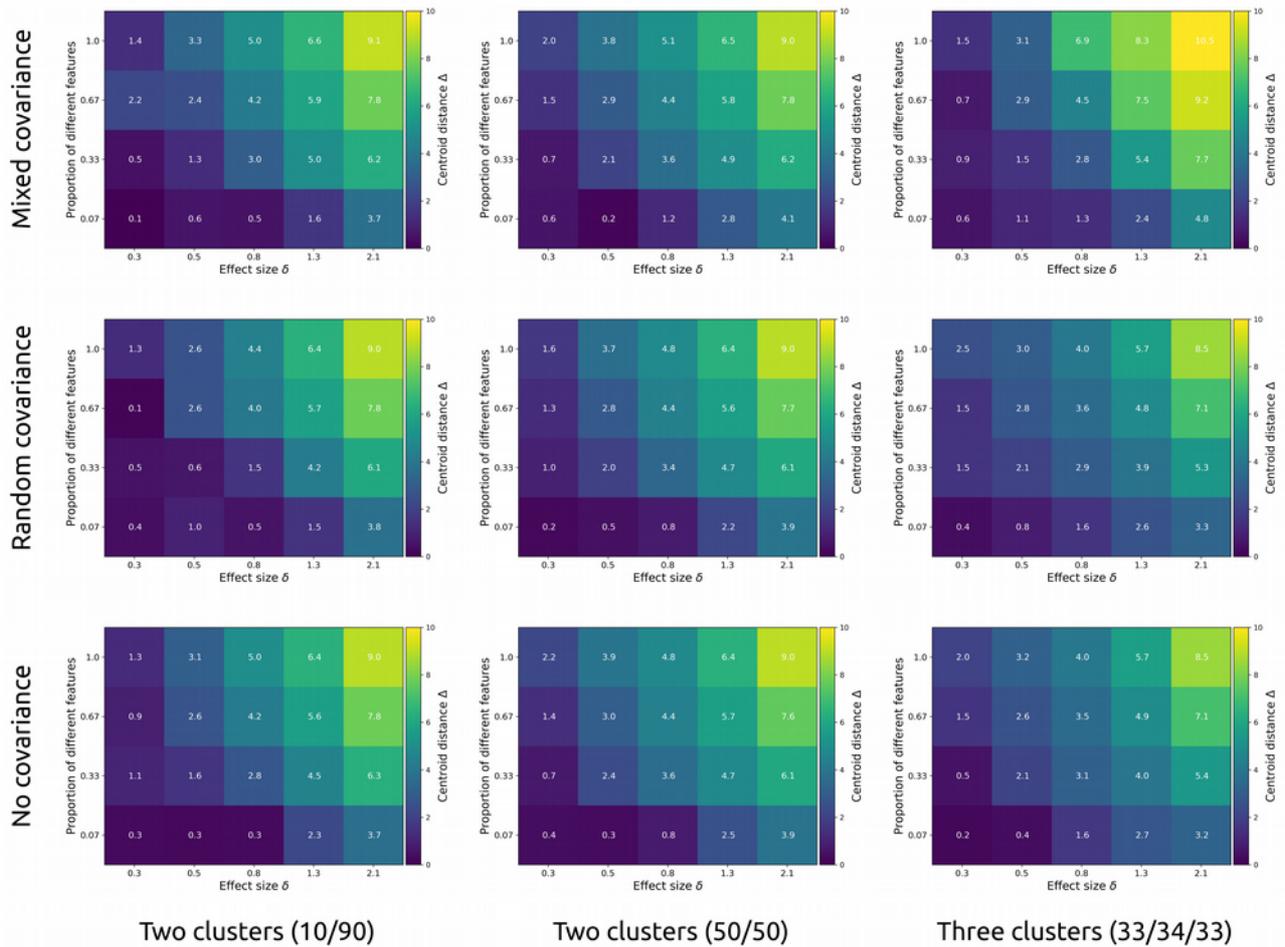

*Figure 3* – Each cell presents the cluster centroid separation Δ (brighter colours indicate stronger separation) after multi-dimensional scaling (MDS) was applied to simulated data of 1000 observations and 15 features. Separation is shown as a function of within-feature effect size (Cohen's d, x-axis), and the proportion of features that were different between clusters. Each row shows a different covariance structure: "mixed" indicates subgroups with different covariance structures, "random" with the same random covariance structure between all groups, and "no" for no correlation between any of the features). Each column shows a different type of population: with unequal (10 and 90%) subgroups, with two equally sized subgroups, and with three equally sized subgroups.

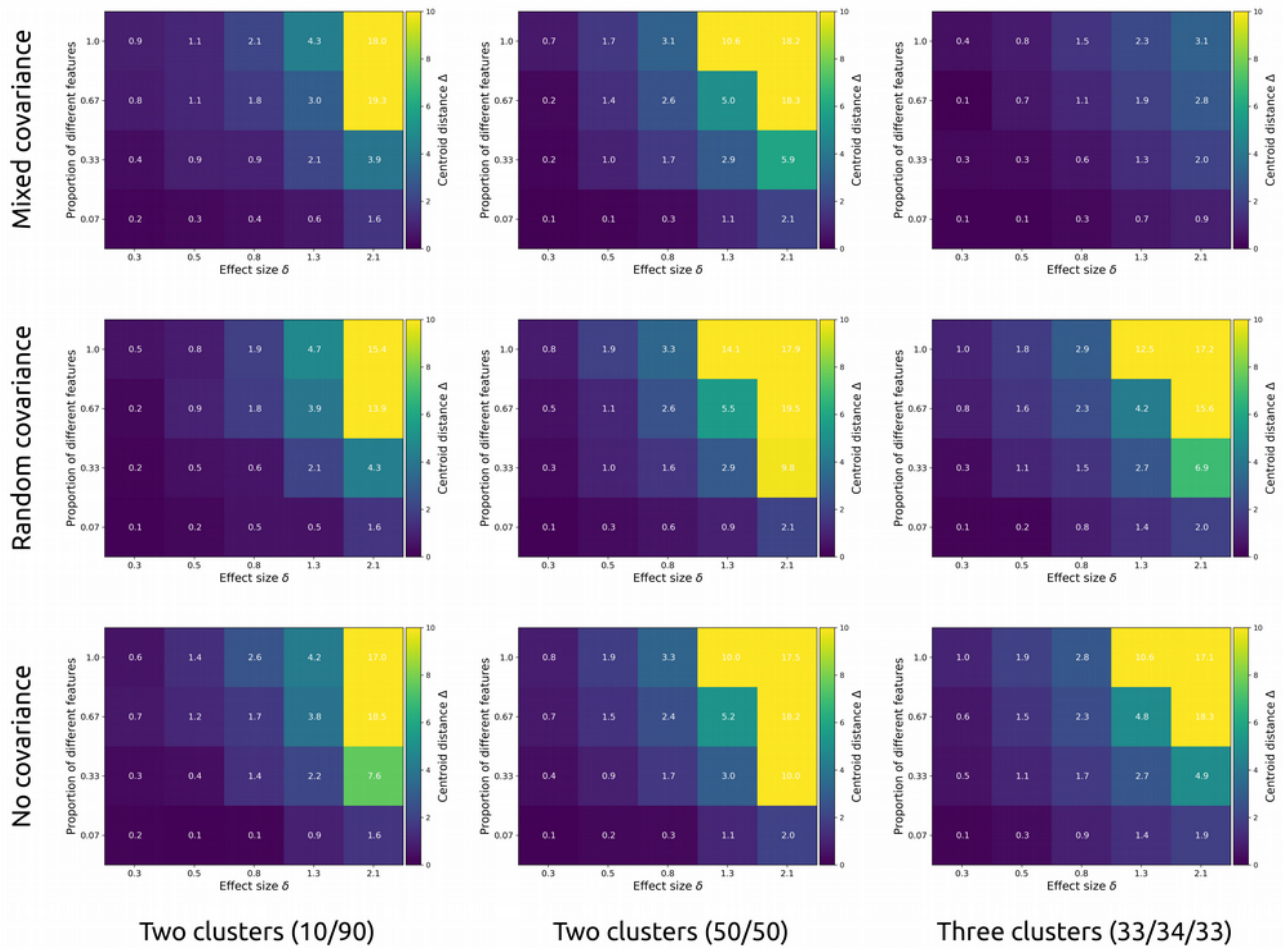

*Figure 4* – Each cell presents the cluster centroid separation Δ (brighter colours indicate stronger separation) after uniform manifold approximation and projection (UMAP) was applied to simulated data of 1000 observations and 15 features. Separation is shown as a function of within-feature effect size (Cohen's d, x-axis), and the proportion of features that were different between clusters. Each row shows a different covariance structure: "mixed" indicates subgroups with different covariance structures, "random" with the same random covariance structure between all groups, and "no" for no correlation between any of the features). Each column shows a different type of population: with unequal (10 and 90%) subgroups, with two equally sized subgroups, and with three equally sized subgroups.

# Membership classification (adjusted Rand index)

Overlap between clustering outcomes and ground truth was computed with the adjusted Rand index. Figure 5 shows the effects of within-feature effect size, number of different features, dimensionality reduction algorithm, and cluster algorithm on the adjusted Rand index for simulated datasets with two equally sized subgroups with different covariance structures (3-factor and 4-factor). These simulated datasets were the optimal example due to their differentiation into two equally sized groups, and their realistic difference in covariance structure.

In terms of clustering accuracy, the k-means algorithm performs roughly equally well regardless of dimensionality reduction. The same is true for the two versions of agglomerative clustering (Ward linkage with Euclidean distance, or average linkage with cosine distance). By contrast, HDBSCAN performs well only after UMAP dimensionality reduction. It is likely that this is due to the algorithm only assigning the denser centres to their respective clusters, while leaving many other observations unassigned.

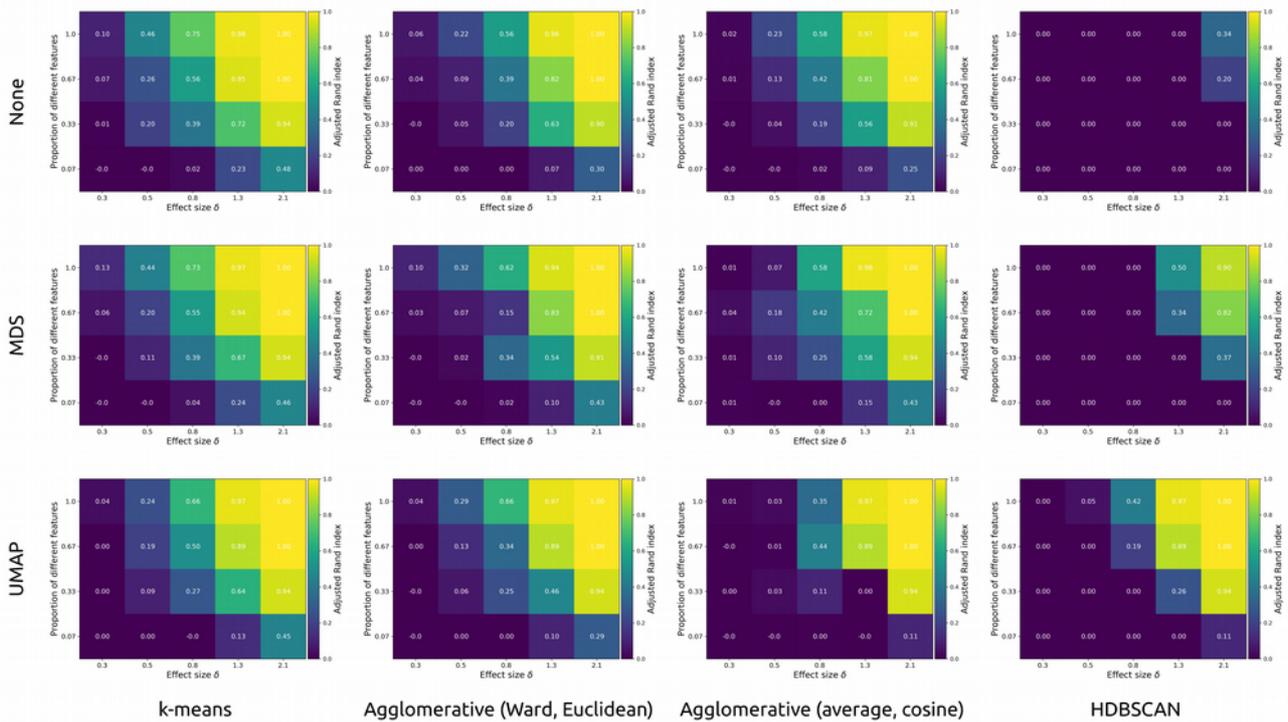

*Figure 5* – Each cell shows the adjusted Rand index (brighter colours indicate better classification) as a function of within-feature effect size (Cohen's d), and the proportion of features that differed between two simulated clusters with different covariance structures (3-factor and 4-factor). Each row presents a different dimensionality reduction approach: None, multi-dimensional scaling (MDS), or uniform manifold approximation and projection (UMAP). Each column presents a different type of clustering algorithm: k-means, agglomerative (hierarchical) clustering with Ward linkage and Euclidean distance, agglomerative clustering with average linkage and cosine distance, and HDBSCAN.

## Data subgrouping (silhouette coefficient)

Silhouette scores reflect the information that would have been available to a researcher to decide on whether discrete subgroups were present in the simulated datasets (i.e. without knowing the ground truth). Figure 6 shows the effects of within-feature effect size, number of different features, dimensionality reduction algorithm, and cluster algorithm on the silhouette scores for simulated datasets two equally sized subgroups with different covariance structures (3-factor and 4-factor). Unlike the adjusted Rand index, for which the ground truth needs to be known, silhouette scores are impacted by dimensionality reduction. Using the traditional threshold of 0.5, none of the raw datasets would have been correctly identified as clustered.

After MDS, only the datasets in which two-thirds or more of the features showed a within-feature difference with a Cohen's *d* of 2.1 would have been correctly identified as showing clustering through k-means and the two agglomerative clustering approaches. On the basis of HDBSCAN, a researcher would have correctly identified clustering in datasets with two-thirds or more features showing a difference with a Cohen's *d* of 1.3, or one-third or more features showing a difference with a Cohen's *d* of 2.1.

Performance is best after UMAP dimensionality reduction. Using k-means, a researcher would correctly identify the clustered nature of datasets with two-thirds of features showing differences of 0.8, or one-third or more showing differences of 1.3 or over. The same is true for HDBSCAN, which was also able to identify clustering when all features showed a within-feature difference corresponding to Cohen's *d* values of 0.3.

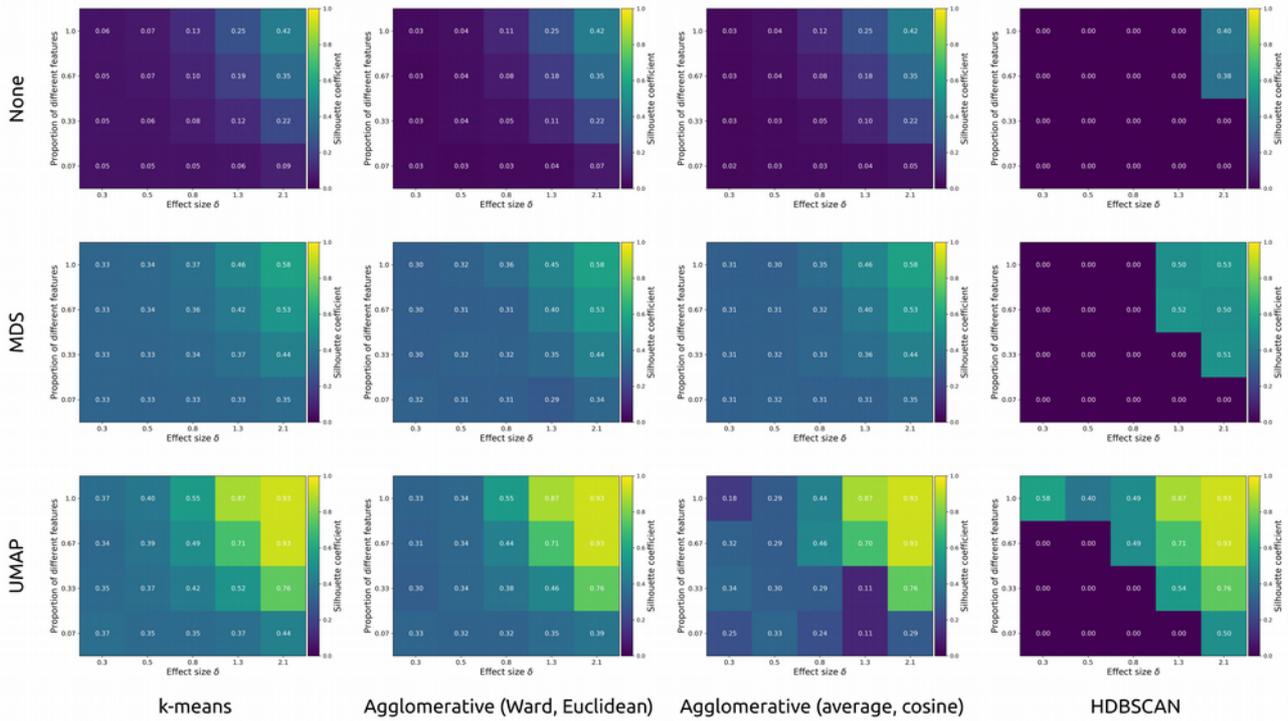

*Figure 6* – Each cell shows the silhouette coefficient (brighter colours indicate stronger detected clustering, with a threshold set at 0.5) as a function of within-feature effect size (Cohen's d), and the proportion of features that differed between two simulated clusters with different covariance structures (3-factor and 4-factor). Each row presents a different dimensionality reduction approach: None, multi-dimensional scaling (MDS), or uniform manifold approximation and projection (UMAP). Each column presents a different type of clustering algorithm: k-means, agglomerative (hierarchical) clustering with Ward linkage and Euclidean distance, agglomerative clustering with average linkage and cosine distance, and HDBSCAN.

In sum, only a minority of the simulated datasets could correctly be identified as clustered, regardless of method. Dimensionality reduction, particularly the non-linear projection provided by UMAP, helped elevate cluster coefficients across the board. Only datasets with large within-feature effect sizes and a high number of features showing differences were correctly marked as "clustered" using the traditional silhouette score threshold of 0.5.

## Effect of dimensionality reduction on cluster separation

As expected, simulated sample centroid distances (Figure 7, in green) were aligned with subgroup centroids before dimensionality reduction was applied, with minor random sampling error. Dimensionality reduction did impact cluster separation. MDS subtly exaggerated centroid distances across all centroid separations in original space (Figure 7, in blue). UMAP reduced sample centroid distance at lower ($\Delta < 3$) and increased it at higher ($\Delta > 4$) centroid distances (Figure 7, in purple). The only exception to this was the simulated dataset with three clusters of respectively 3-factor, 4-factor, and no covariance structure, where UMAP reduced centroid distance for all original centroid distances (it is unclear why, and likely due to chance).

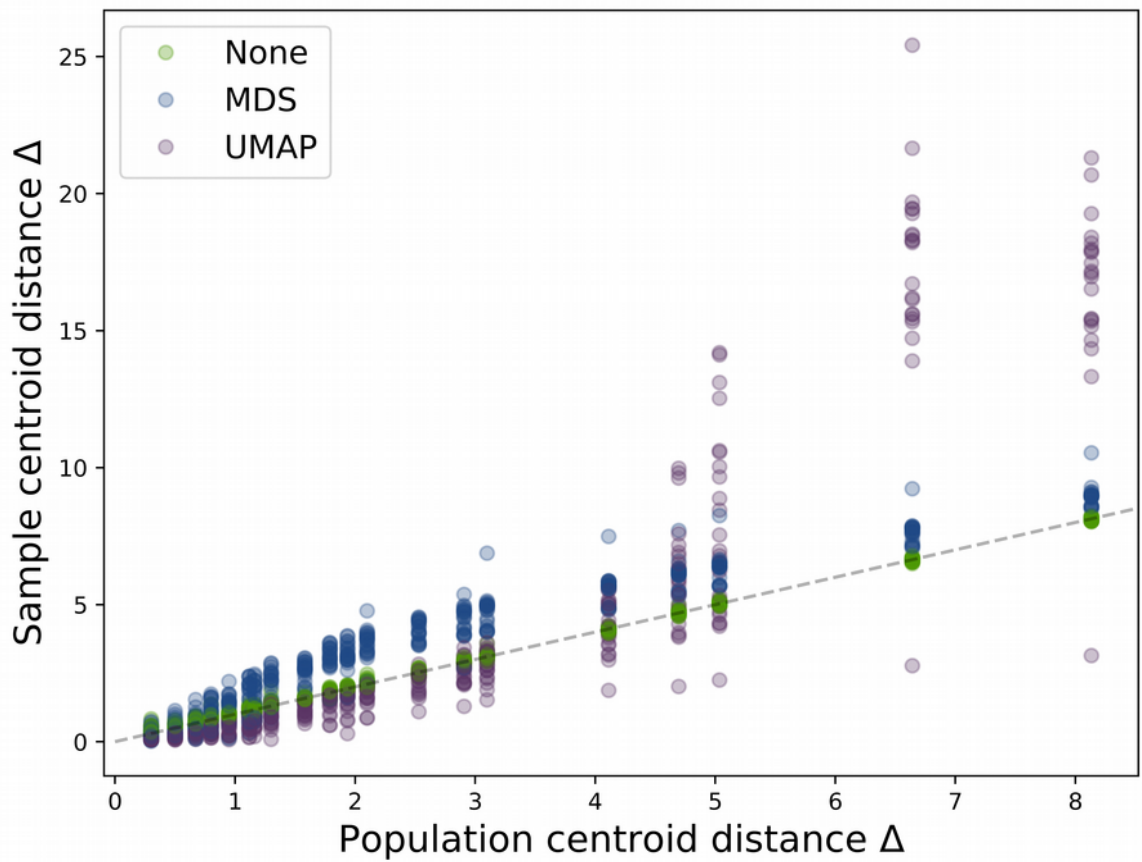

*Figure 7* – Each dot represents a simulated dataset of 1000 observations and 15 features, each with different within-feature effect size, proportion of different features between clusters, covariance structure, and number of clusters. The x-axis represents the separation between subgroups in the population that datasets were simulated from, and the y-axis represents the separation in the simulated dataset after no dimensionality reduction (green), multi-dimensional scaling (MDS, blue), or uniform manifold approximation and projection (purple). The dotted line indicates no difference between population and sample; values above the dotted line have had their separation increased, and it was decreased for those below the line.

In sum, subgroups were separated to different extents in reduced and original space. MDS was likely to increase separation, and UMAP was likely to increase it dramatically when the original separation was large enough. Due to the stochastic nature of the algorithms, the size of their effects on separation is variable.

## Statistical power and accuracy

We opted to not present the outcomes of agglomerative clustering, because they align closely with k-means. In addition, because the previous analyses showed that subgroup centroid separation was the main factor in determining cluster analysis outcomes, we opted for simulating datasets in reduced space (multivariate normal with two features).

Power was computed as the probability of correctly rejecting the null hypothesis of there not being a clustered structure to the data, which occurred when silhouette scores were 0.5 or over. The probability of selecting the correct number of clusters was computed as the proportion of solutions where the silhouette score was highest for the true number of clusters. Finally, classification accuracy was computed as the proportion of observations that was correctly assigned to their respective cluster.

### K-means

For k-means, power to detect clustering was primarily dependent on cluster separation, and much less on sample size (Table 2; Figure 8, top and second row). At cluster separation $\Delta=5$, there was 71% power to detect clustering in a population divided into one large (90%) and one small (10%) subgroup at sample size N=10, and 92% at N=20. For two equally sized clusters, power was 82% from separation $\Delta=4$ at N=10, and higher for larger sample and effect sizes. For three equally sized clusters, power was 76% at separation $\Delta=4$ for N=10, 69% for N=20, 77% to N=40, and over 80 from N=80; with power for larger effect sizes around 100%. For four equally sized clusters, power was 75% at separation Δ=4 at N=80, 85% at Δ=5 for N=10, and around 100% for larger effect and sample sizes. (See Figure 8 and Table 2 for all quoted numbers.)

Sample sizes of N=40 resulted in good (80% or higher) accuracy to detect the true number of clusters from separation Δ=4. For equally sized clusters, that level of accuracy was also reached at separation Δ=3 when the sample size was about 20 per subgroup (Figure 8, third row).

Classification accuracy for subgroup membership of individual observations was above chance for all tested separation values and sample sizes in populations with equally sized subgroups, and above 80-90% from separation Δ=3. Classification accuracy was above chance for populations with one small (10%) and one large (90%) subgroup from N=80 and separation Δ=4 (Figure 8, bottom row).

In sum, 20 observations per subgroup resulted in sufficient power to detect the presence of subgroups with k-means, provided cluster separation was Δ=4 or over, and subgroups were roughly equally sized (detecting smaller subgroups among large subgroups was only possible for separations of Δ=5 or over). These values also provided near-perfect accuracy for the detection of the true number of clusters, and very high (90-100%) classification accuracy of individual observation's group membership.

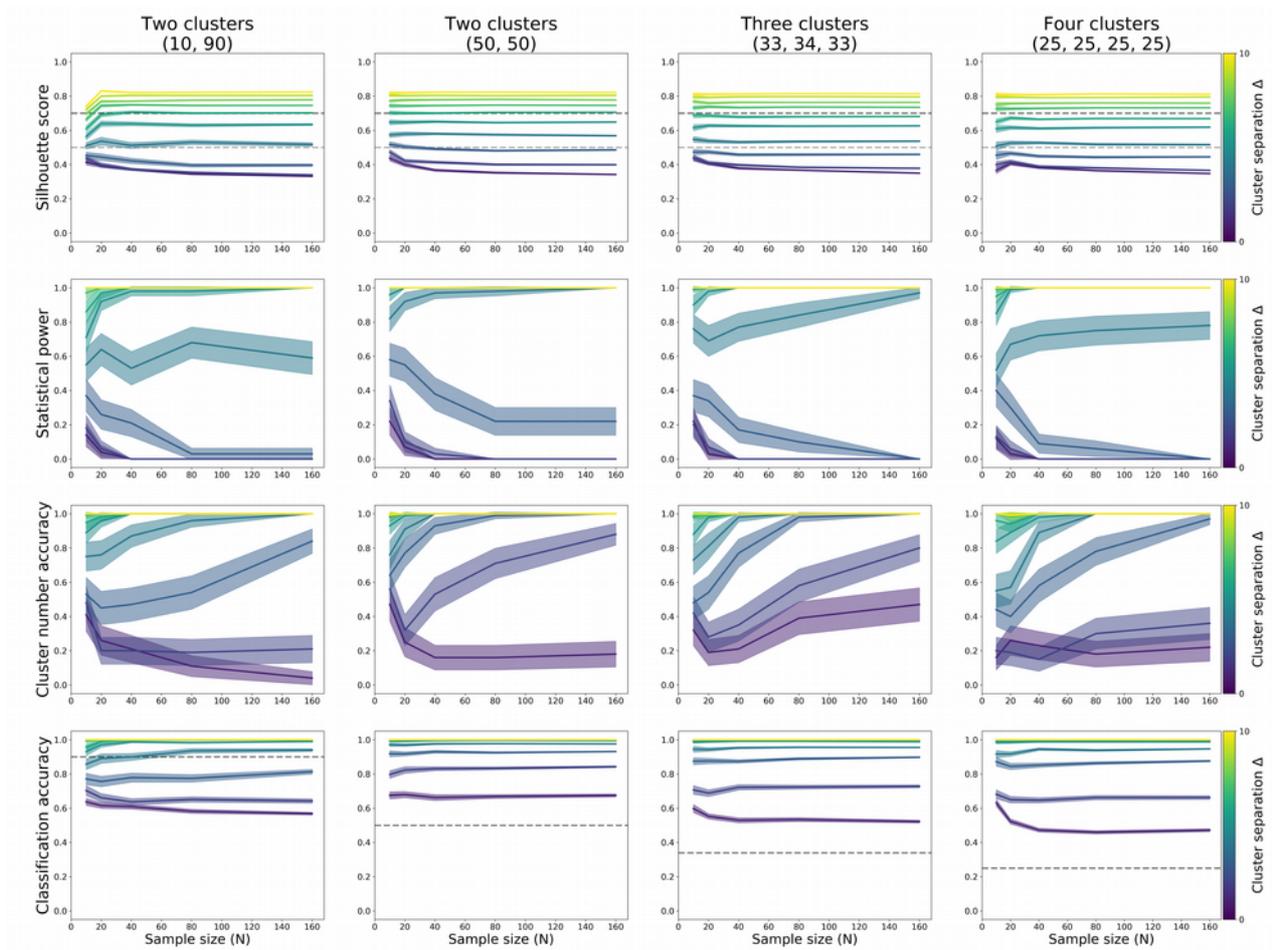

*Figure 8* – K-means silhouette scores (top row), proportion of correctly identified clustering (second row), proportion of correctly identified number of clusters (third row), and the proportion of observations correctly assigned to their subgroup, each computed through 100 iterations of simulation. Datasets of varying sample size (x-axis) and two features were sampled from populations with equidistant subgroups that each had the same 3-factor covariance structure. The simulated populations were made up of two unequally sized (10 and 90%) subgroups (left column); or two, three, or four equally sized subgroups (second, third, and fourth column, respectively).

**Table 2**

*Statistical power for the binary decision of data being "clustered" using k-means clustering.*

*Estimates based on 100 iterations per cell, using a decision threshold of 0.5 for silhouette scores.*

|  | Δ=1 | Δ=2 | Δ=3 | Δ=4 | Δ=5 | Δ=6 | Δ=7 | Δ=8 | Δ=9 | Δ=10 |
|---|---|---|---|---|---|---|---|---|---|---|
| **2 Clusters** | | | | | | | | | | |
| (10/90%) | | | | | | | | | | |
| N=10 | 14 | 18 | 37 | 55 | 71 | 86 | 97 | 100 | 100 | 100 |
| N=20 | 4 | 6 | 26 | 64 | 92 | 97 | 100 | 100 | 100 | 100 |
| N=40 | 0 | 0 | 21 | 53 | 98 | 100 | 100 | 100 | 100 | 100 |
| N=80 | 0 | 0 | 3 | 68 | 98 | 100 | 100 | 100 | 100 | 100 |
| N=160 | 0 | 0 | 3 | 59 | 100 | 100 | 100 | 100 | 100 | 100 |
| **2 Clusters** | | | | | | | | | | |
| (50/50%) | | | | | | | | | | |
| N=10 | 22 | 34 | 58 | 82 | 96 | 100 | 100 | 100 | 100 | 100 |
| N=20 | 7 | 10 | 55 | 92 | 100 | 100 | 100 | 100 | 100 | 100 |
| N=40 | 0 | 3 | 38 | 97 | 100 | 100 | 100 | 100 | 100 | 100 |
| N=80 | 0 | 0 | 22 | 98 | 100 | 100 | 100 | 100 | 100 | 100 |
| N=160 | 0 | 0 | 22 | 100 | 100 | 100 | 100 | 100 | 100 | 100 |
| **3 Clusters** | | | | | | | | | | |
| (33/34/33%) | | | | | | | | | | |
| N=10 | 22 | 20 | 37 | 76 | 90 | 99 | 100 | 100 | 100 | 100 |
| N=20 | 3 | 7 | 34 | 69 | 98 | 100 | 100 | 100 | 100 | 100 |
| N=40 | 0 | 0 | 17 | 77 | 100 | 100 | 100 | 100 | 100 | 100 |
| N=80 | 0 | 0 | 10 | 84 | 100 | 100 | 100 | 100 | 100 | 100 |
| N=160 | 0 | 0 | 0 | 97 | 100 | 100 | 100 | 100 | 100 | 100 |
| **4 Clusters** | | | | | | | | | | |
| (25/25/25/25%) | | | | | | | | | | |
| N=10 | 12 | 13 | 40 | 52 | 85 | 95 | 99 | 100 | 100 | 100 |
| N=20 | 3 | 6 | 30 | 67 | 99 | 100 | 100 | 100 | 100 | 100 |
| N=40 | 0 | 0 | 9 | 72 | 100 | 100 | 100 | 100 | 100 | 100 |
| N=80 | 0 | 0 | 6 | 75 | 100 | 100 | 100 | 100 | 100 | 100 |
| N=160 | 0 | 0 | 0 | 78 | 100 | 100 | 100 | 100 | 100 | 100 |

## HDBSCAN

For HDBSCAN, power to detect clustering was primarily dependent on effect size, provided the sample size was over a threshold (Table 3; Figure 9, top and second row). For a population divided into one large (90%) and one small (10%) cluster, power was 84% at N=80 for separation Δ=6. For two clusters of equal size, power was 66% at N=40 and 83% for N=80 for separation Δ=3. For three clusters of equal size, power was 66% at N=160 for separation Δ=3, and 84% at N=80 for separation Δ=4. For four clusters of equal size, power was 75% at N=80 for separation Δ=4. (See Figure 9 and Table 3 for all quoted numbers.)

In addition to detecting subgrouping in the data, Figure 9 (third row) shows the probability that HDBSCAN detected the correct number of subgroups. This, too, was strongly dependent on cluster separation. Accuracy was highest for separations of Δ=5 or over, and mostly acceptable (around 70-80%) at separations Δ=4. For equally sized clusters, this was true from N=40, while the accurate detection of one small (10%) and one large (90%) subgroup required N=80 or over.

The accuracy of classifying observations' cluster membership (Figure 9, bottom row) was only over chance at from separations of Δ=8 and sample sizes of N=80 for populations with one small (10%) and one large (90%) subgroup (chance = 90%). It was over chance from Δ=4 and N=40 for populations with two equally sized subgroups (chance = 50%), from Δ=4 and N=40 for populations with three equally sized subgroups (chance = 33%), and from Δ=3 and N=40 for populations with four equally sized subgroups (chance = 25%).

In sum, 20-30 observations per subgroup resulted in sufficient power to detect the presence of subgroups with HDBSCAN, provided cluster separation was Δ=4 or over, and subgroups were roughly equally sized (detecting smaller subgroups among large subgroups was only possible for separations of Δ=6 or over). These values also provided reasonable accuracy for the detection of the true number of clusters, as well as over-chance classification accuracy of individual observation's group membership.

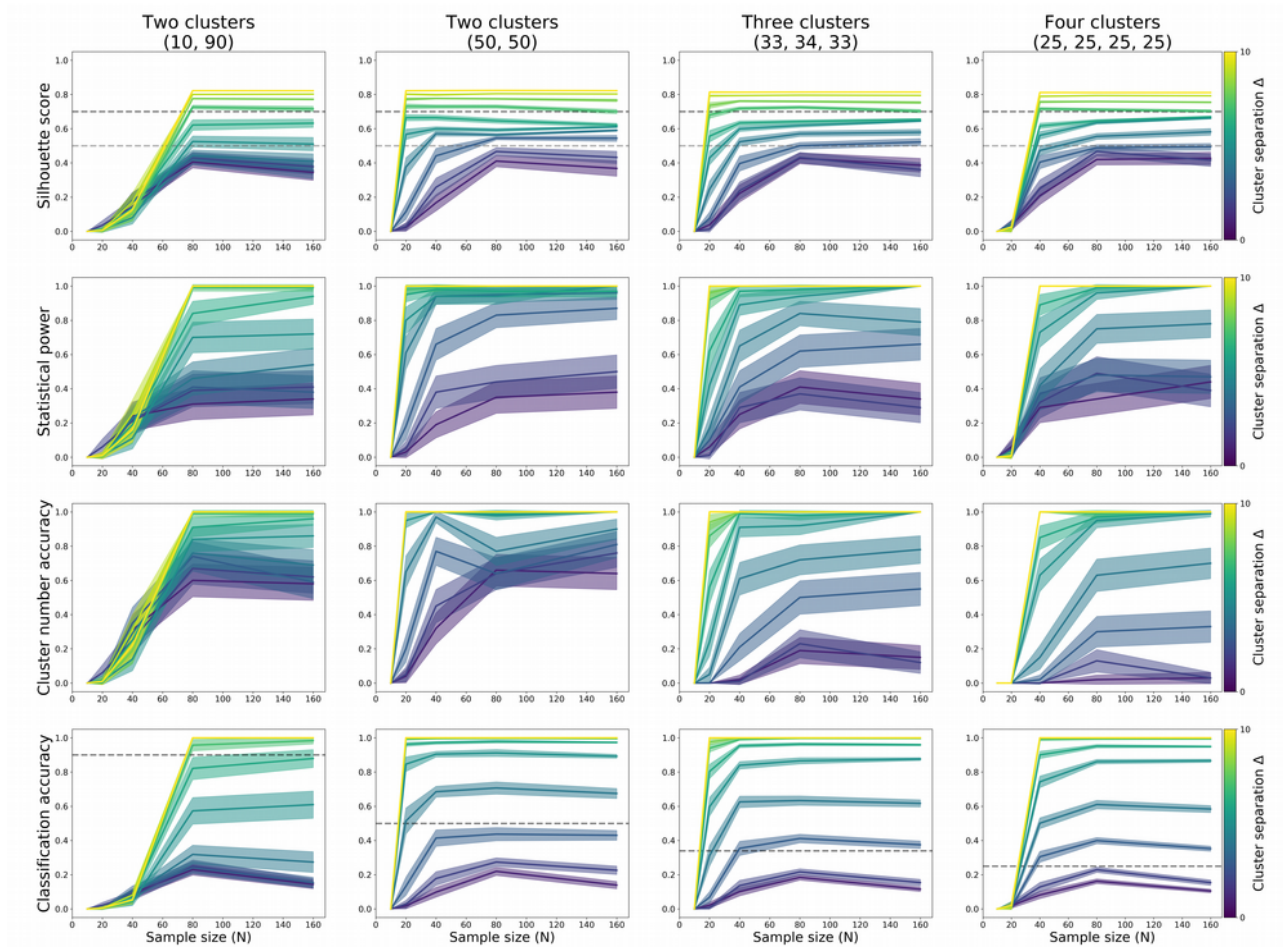

*Figure 9* – HDBSCAN silhouette scores (top row), proportion of correctly identified clustering (second row), proportion of correctly identified number of clusters (third row), and the proportion of observations correctly assigned to their subgroup, each computed through 100 iterations of simulation. Datasets of varying sample size (x-axis) and two features were sampled from populations with equidistant subgroups that each had the same 3-factor covariance structure. The simulated populations were made up of two unequally sized (10 and 90%) subgroups (left column); or two, three, or four equally sized subgroups (second, third, and fourth column, respectively).

**Table 3**

*Statistical power for the binary decision of data being "clustered" using HDBSCAN clustering.*

*Estimates based on 100 iterations per cell, using a decision threshold of 0.5 for silhouette scores.*

|  | Δ=1 | Δ=2 | Δ=3 | Δ=4 | Δ=5 | Δ=6 | Δ=7 | Δ=8 | Δ=9 | Δ=10 |
|---|---|---|---|---|---|---|---|---|---|---|
| **2 Clusters** | | | | | | | | | | |
| (10/90%) | | | | | | | | | | |
| **N=10** | 0 | 0 | 0 | 0 | 0 | 0 | 0 | 0 | 0 | 0 |
| **N=20** | 0 | 6 | 0 | 1 | 1 | 1 | 0 | 0 | 1 | 1 |
| **N=40** | 24 | 21 | 22 | 23 | 11 | 19 | 15 | 16 | 25 | 16 |
| **N=80** | 31 | 39 | 41 | 46 | 70 | 84 | 99 | 100 | 100 | 100 |
| **N=160** | 34 | 41 | 38 | 54 | 72 | 94 | 99 | 100 | 100 | 100 |
| **2 Clusters** | | | | | | | | | | |
| (50/50%) | | | | | | | | | | |
| **N=10** | 0 | 0 | 0 | 0 | 0 | 0 | 0 | 0 | 0 | 0 |
| **N=20** | 3 | 5 | 18 | 61 | 80 | 95 | 98 | 100 | 100 | 100 |
| **N=40** | 19 | 38 | 66 | 94 | 94 | 98 | 99 | 100 | 100 | 100 |
| **N=80** | 35 | 44 | 83 | 95 | 94 | 97 | 100 | 100 | 100 | 100 |
| **N=160** | 38 | 50 | 87 | 96 | 100 | 97 | 100 | 100 | 100 | 100 |
| **3 Clusters** | | | | | | | | | | |
| (33/34/33%) | | | | | | | | | | |
| **N=10** | 0 | 0 | 0 | 0 | 0 | 0 | 0 | 0 | 0 | 0 |
| **N=20** | 6 | 1 | 10 | 17 | 42 | 62 | 92 | 96 | 100 | 100 |
| **N=40** | 25 | 29 | 41 | 65 | 89 | 97 | 100 | 100 | 100 | 100 |
| **N=80** | 41 | 37 | 62 | 84 | 94 | 98 | 100 | 100 | 100 | 100 |
| **N=160** | 34 | 29 | 66 | 79 | 100 | 100 | 100 | 100 | 100 | 100 |
| **4 Clusters** | | | | | | | | | | |
| (25/25/25/25%) | | | | | | | | | | |
| **N=10** | 0 | 0 | 0 | 0 | 0 | 0 | 0 | 0 | 0 | 0 |
| **N=20** | 6 | 5 | 2 | 2 | 2 | 2 | 0 | 0 | 0 | 1 |
| **N=40** | 29 | 32 | 37 | 42 | 73 | 89 | 100 | 100 | 100 | 100 |
| **N=80** | 34 | 49 | 48 | 75 | 96 | 99 | 100 | 100 | 100 | 100 |
| **N=160** | 44 | 39 | 47 | 78 | 100 | 100 | 100 | 100 | 100 | 100 |

## C-means

As for k-means and HDBSCAN, for c-means power to detect clustering was primarily dependent on cluster separation, and much less on sample size (Table 4; Figure 10, top and second row). At cluster separation Δ=5, there was 81% power to detect clustering in a population divided into one large (90%) and one small (10%) subgroup at sample size N=20, and 95-100% at larger effect and sample sizes. For two equally sized clusters, power was 77% for N=10 and 82% and N=20 for separation Δ=3, and 91-100% for larger sample and effect sizes. For three equally sized clusters, power was 77% at separation Δ=3 for N=40, 76% for N=10 at Δ=4, and 89-100% for larger sample and effect sizes. For four equally sized clusters, power was 75% for N=40 and 83% at N=80 at separation Δ=3, 94% for N=20 at Δ=4, and around 100% for larger effect and sample sizes. (See Figure 10 and Table 4 for all quoted numbers.)

Sample sizes of N=40 resulted in good (80% or higher) accuracy to detect the true number of clusters from separation Δ=4. For equally sized clusters, that level of accuracy was also reached at separation Δ=3 when the sample size was about 20 per subgroup (Figure 10, third row).

Classification accuracy for subgroup membership of individual observations was above chance for all tested separation values and sample sizes in populations with equally sized subgroups, and above 80-90% from separation Δ=3. Classification accuracy was above chance for populations with one small (10%) and one large (90%) subgroup from N=40 and separation Δ=5 (Figure 10, bottom row).

In sum, 20 observations per subgroup resulted in sufficient power to detect the presence of subgroups with c-means, provided cluster separation was Δ=3 or over, and subgroups were roughly equally sized (detecting smaller subgroups among large subgroups was only possible for separations of Δ=5 or over). These values also provided near-perfect accuracy for the detection of the true number of clusters, and very high (90-100%) classification accuracy of individual observation's group membership.

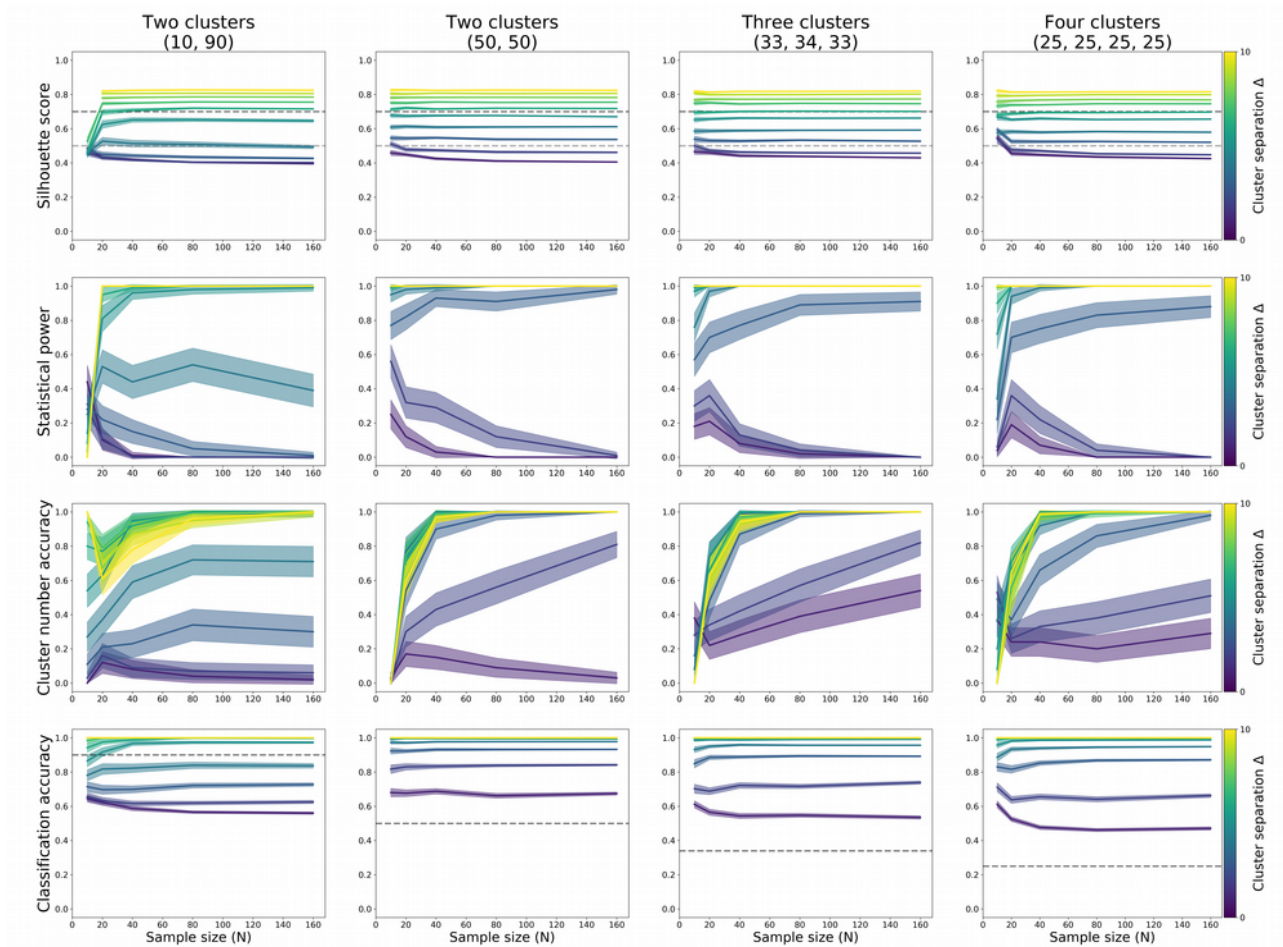

*Figure 10* – *C-means silhouette scores (top row), proportion of correctly identified clustering (second row), proportion of correctly identified number of clusters (third row), and the proportion of observations correctly assigned to their subgroup, each computed through 100 iterations of simulation. Datasets of varying sample size (x-axis) and two features were sampled from populations with equidistant subgroups that each had the same 3-factor covariance structure. The simulated populations were made up of two unequally sized (10 and 90%) subgroups (left column); or two, three, or four equally sized subgroups (second, third, and fourth column, respectively).*

**Table 4**

*Statistical power for the binary decision of data being "clustered" using k-means clustering.*

*Estimates based on 100 iterations per cell, using a decision threshold of 0.5 for silhouette scores.*

|  | Δ=1 | Δ=2 | Δ=3 | Δ=4 | Δ=5 | Δ=6 | Δ=7 | Δ=8 | Δ=9 | Δ=10 |
|---|---|---|---|---|---|---|---|---|---|---|
| **2 Clusters** | | | | | | | | | | |
| (10/90%) | | | | | | | | | | |
| N=10 | 44 | 28 | 31 | 25 | 14 | 8 | 3 | 0 | 0 | 0 |
| N=20 | 10 | 11 | 22 | 53 | 81 | 95 | 100 | 100 | 100 | 100 |
| N=40 | 1 | 0 | 15 | 44 | 96 | 99 | 100 | 100 | 100 | 100 |
| N=80 | 0 | 0 | 5 | 54 | 98 | 100 | 100 | 100 | 100 | 100 |
| N=160 | 0 | 0 | 1 | 39 | 99 | 100 | 100 | 100 | 100 | 100 |
| **2 Clusters** | | | | | | | | | | |
| (50/50%) | | | | | | | | | | |
| N=10 | 25 | 56 | 77 | 95 | 100 | 99 | 100 | 100 | 100 | 100 |
| N=20 | 12 | 32 | 82 | 98 | 100 | 100 | 100 | 100 | 100 | 100 |
| N=40 | 3 | 29 | 93 | 99 | 100 | 100 | 100 | 100 | 100 | 100 |
| N=80 | 0 | 12 | 91 | 100 | 100 | 100 | 100 | 100 | 100 | 100 |
| N=160 | 0 | 1 | 98 | 100 | 100 | 100 | 100 | 100 | 100 | 100 |
| **3 Clusters** | | | | | | | | | | |
| (33/34/33%) | | | | | | | | | | |
| N=10 | 18 | 30 | 57 | 76 | 97 | 99 | 100 | 100 | 100 | 100 |
| N=20 | 21 | 36 | 70 | 97 | 100 | 100 | 100 | 100 | 100 | 100 |
| N=40 | 8 | 13 | 77 | 100 | 100 | 100 | 100 | 100 | 100 | 100 |
| N=80 | 2 | 4 | 89 | 100 | 100 | 100 | 100 | 100 | 100 | 100 |
| N=160 | 0 | 0 | 91 | 100 | 100 | 100 | 100 | 100 | 100 | 100 |
| **4 Clusters** | | | | | | | | | | |
| (25/25/25/25%) | | | | | | | | | | |
| N=10 | 4 | 6 | 22 | 34 | 72 | 90 | 100 | 99 | 100 | 100 |
| N=20 | 19 | 36 | 70 | 94 | 100 | 100 | 100 | 100 | 100 | 100 |
| N=40 | 7 | 22 | 75 | 99 | 100 | 100 | 100 | 100 | 100 | 100 |
| N=80 | 0 | 4 | 83 | 100 | 100 | 100 | 100 | 100 | 100 | 100 |
| N=160 | 0 | 0 | 88 | 100 | 100 | 100 | 100 | 100 | 100 | 100 |

# Direct comparison of discrete and fuzzy clustering

The power results summarised above suggested c-means (80-100% power at Δ=3) is more powerful than k-means (70-100% power at Δ=4) for detecting equally sized clusters. K-means outcomes were interpreted using the traditional silhouette score, whereas c-means outcomes were using the fuzzy silhouette score. It could be that the latter inflated silhouette scores, which would make c-means more likely to detect clustering at lower centroid separations (see Figure 11 for an example between Δ=3 and 4). However, such an inflation could have also increased the likelihood of false positives.

To test this, we simulated data without clustering (k=1), and data with equidistant clusters of equal size (k=2 to 4) with a range of centroid separations (Δ=1 to 10). In addition to c-means, we also employed Gaussian mixture modelling. This is a different approach than c-means clustering, but it does result in a similar metric of membership confidence for each observation. This was used to compute the fuzzy silhouette coefficient in the same way as for c-means.

Fuzzy silhouette scores were indeed higher than traditional silhouette scores (Figure 12). Crucially, the false positive rate (the probability of silhouette scores to surpass the 0.5 threshold when no clustering was present) was 0% for k-means, c-means, and Gaussian mixture modelling (Figure 12, left column; based on 100 iterations, each algorithm used on the same simulated data). In sum, using non-discrete methods and the fuzzy cluster coefficient increased the likelihood of cluster detection, but not the false positive rate.

Figure 13 shows power (proportion of iterations that found "clustering", i.e. a silhouette score of 0.5 of higher) and cluster number accuracy (proportion of iterations in which the ground truth number of clusters was correctly identified) as a function of cluster separation (effect size) in simulated data. It confirms that c-means and Gaussian mixture modelling (100% power at Δ=3) are more powerful than k-means (100% at Δ=4). C-means and mixture modelling also succeeded at identifying the correct number of clusters at lower separations than k-means. Finally, estimating the number of clusters on the basis of the highest silhouette score was a better method than choosing on the basis of the lowest Bayesian Information Criterion (BIC) with Gaussian mixture modelling.

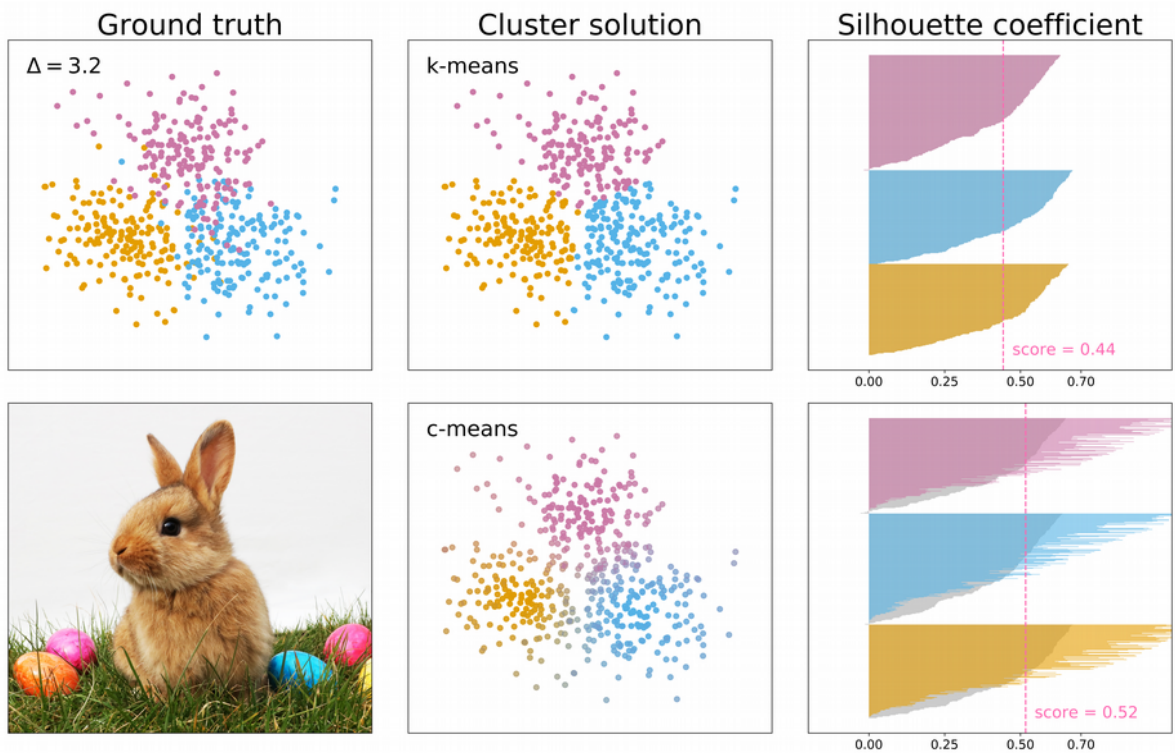

*Figure 11* – *Clustering outcomes of discrete k-means clustering (top row) and fuzzy c-means clustering (bottom row). The top-left panel shows the ground truth, a sample (N=600) from a simulated population made up of three equally sized and equidistant subgroups. The middle column shows the assignment of observations to clusters, and the right column shows the corresponding silhouette coefficients. The shading in the bottom silhouette plot indicates the discrete clustering silhouette coefficients, and the coloured bars indicate a transformation equivalent to that performed to compute the fuzzy silhouette score. The bottom-left shows a bunny (also fuzzy) amid contained and well-separated ellipsoidal clusters (photo by Tim Reckmann, licensed CC-BY 2.0).*

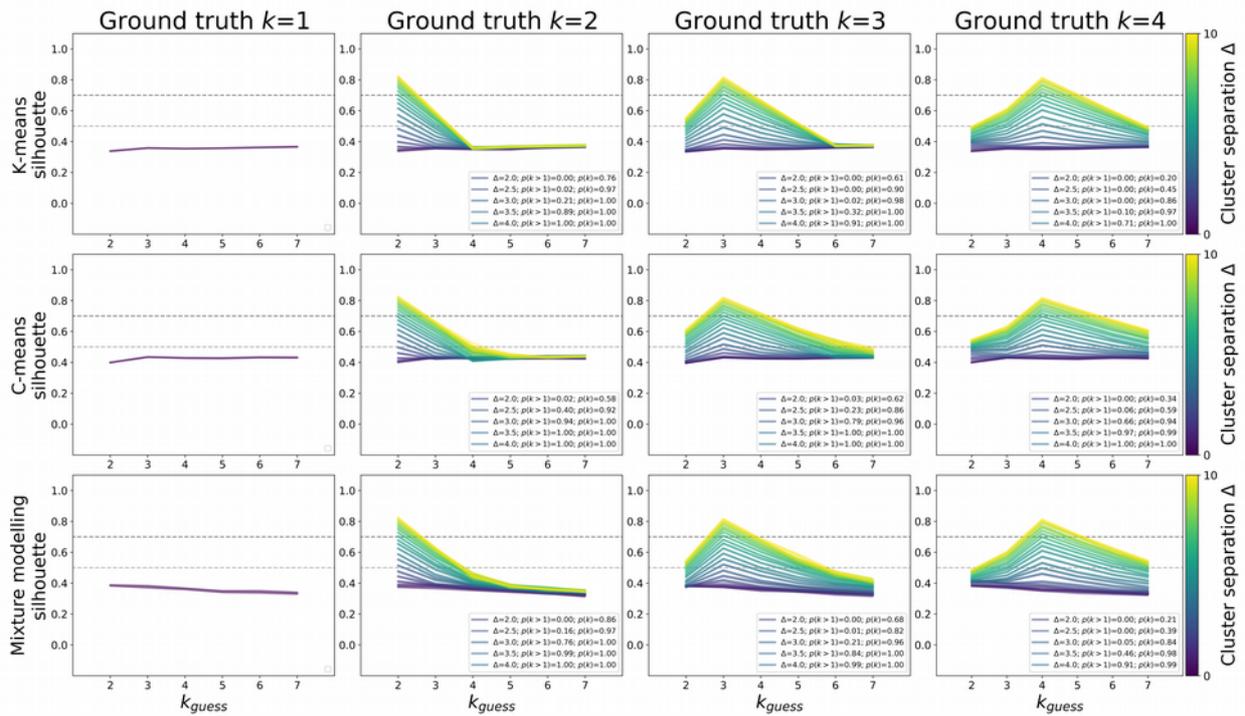

*Figure 12* – *Traditional (top row), fuzzy (middle row and bottom row) silhouette scores, respectively computed after clustering through k-means, c-means, and Gaussian finite mixture modelling. Each line presents the mean and 95% confidence interval obtained over 100 iterations of sampling (N=120, two uncorrelated features) from populations with no subgroups (left column), two subgroups (second column), three subgroups (third column), or four subgroups (right column) that were equidistant and of equal size. The same dataset in each iteration was subjected to k-means, c-means, and Gaussian mixture modelling. Estimates were obtained for different centroid separation values (Δ=1 to 10, in steps of 0.5; brighter colours indicate stronger separation). All simulation results are plotted, and power (proportion of iterations that found k>1) and accuracy (proportion of iterations that found the true number of clusters) are annotated for centroid separations Δ=2 to 4 (see Figure 13 for power and accuracy as a function of centroid separation).*

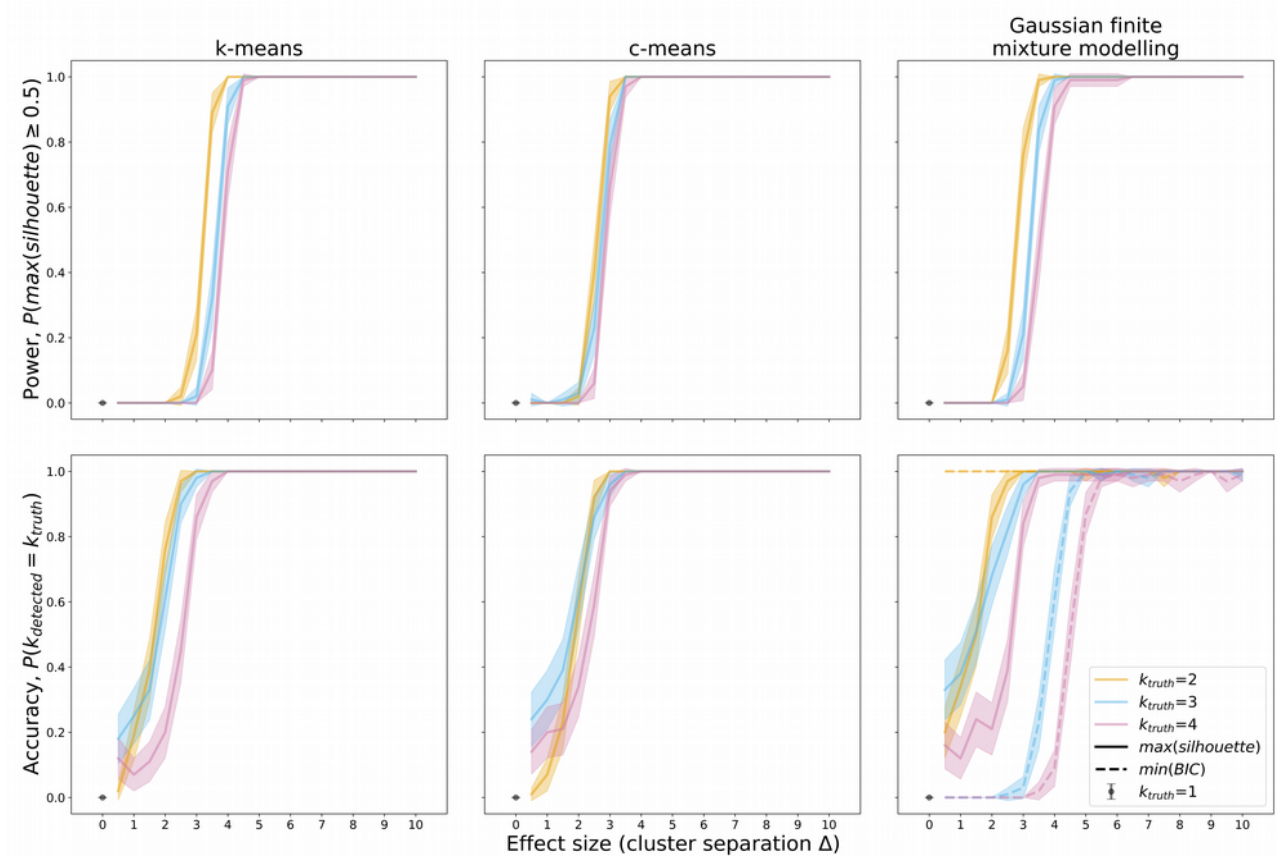

*Figure 13 – Power (top row) and cluster number accuracy (bottom row) for k-means (left column), c-means (middle column), and Gaussian mixture modelling (right column). Power was computed as the proportion of simulations in which the silhouette score was equal to or exceeded 0.5, the threshold for subgroups being present in the data. Cluster number accuracy was computed as the proportion of simulations in which the highest silhouette coefficient (solid lines) or the lowest Bayesian Information Criterion (dashed line, only applicable in Gaussian mixture modelling) was associated with the true simulated number of clusters. For each ground truth k=1 to k=4, 100 simulation iterations were run. In each iteration, all three methods were employed on the same data, and all methods were run for guesses k=2 to k=7.*

# Discussion

Ensuring adequate statistical power is essential to improve reliability and replicability of science (27–29). Furthermore, the decision of whether subgroups exist in data can have important theoretical and clinical consequences, for example when cluster analysis is used as a data-driven approach to define diagnostic subgroups (35), or grouping patients in clinical practice (36). We employed a simulation approach to determining statistical power for cluster analyses, i.e. the probability of correctly accepting the hypothesis that subgroups exist in multivariate data. We simulated subgroups as multivariate normal distributions that varied in number, relative size, separation, and covariance structure. We also varied dimensionality reduction technique (none, multi-dimensional scaling, and uniform manifold approximation and projection) and cluster-detection algorithm (k-means, agglomerative hierarchical clustering, HDBSCAN, c-means, and Gaussian mixture modelling).

We found that covariance structure did not impact cluster analysis outcomes. Dimensionality reduction through multi-dimensional scaling (MDS) increases subgroup separation by about $\Delta=1$, whereas uniform manifold approximation and projection (UMAP) decreases separation when it was below $\Delta=4$ in original space, or increase it when original separation was over $\Delta=5$.

While centroid separation is the main driver of statistical power in cluster analysis, sample size and relative subgroup size had some effect. In populations with a small subgroup (10%), larger separation was required to properly detect clustering. Furthermore, most algorithms performed optimally at lower separations only with a minimum sample size of 20-30 observations per subgroup, and thus a total sample size of that multiplied by the number of expected subgroups (k) within a studied population.

Our results also showed that statistical power is better for "fuzzy" methods compared to traditional discrete methods. Specifically, c-means had sufficient power (80-100%) from a centroid separation of Δ=3, whereas that was only true from Δ=4 for k-means.

Finally, statistical power was similar between c-means and finite Gaussian mixture modelling. This is noteworthy, because latent profile analysis and latent class analysis (both branches of finite mixture modelling) are increasingly popular among researchers (Figure 1). Our results suggest that both fuzzy clustering and finite mixture modelling approaches are more powerful than discrete clustering approaches.

## Take-home messages

- Cluster analyses are not sensitive to covariance structure, or differences in covariance structure between subgroups.

- If you are testing the hypothesis that subgroups with different means (centroids) exist within your population, cluster analysis will only be able to confirm this if the groups show strong separation. You can compute the expected separation in your data using Equation 1.

- Dimensionality reduction through MDS subtly increases separation Δ by about 1 in our simulations, and hence can help improve your odds of accurately detecting clustering. By contrast, dimensionality reduction UMAP will underestimate cluster separation if separation in original feature space is below Δ=4, but it will strongly increase cluster separation for original separation values over 5. Hence, in the context of multivariate normal distributions, we would recommend using MDS.

- Provided subgroups are sufficiently separated in your data (Δ=4), sampling at least N=20-30 observations per group will provide sufficient power to detect subgrouping with k-means or HDBSCAN, with decent accuracy for both the detection of the number of clusters in your sample, and the classification of individual observations' cluster membership.

- Finally, using non-discrete methods (c-means or finite mixture modelling) and the fuzzy silhouette score could improve the power of cluster analyses without elevating the false positive rate. This is particularly useful if the expected cluster centroid separation is between Δ=3 to 4, where k-means is less likely to reliably detect subgroups than fuzzy alternatives.

## Practical example

This example summarises the above take-home measures in a hypothetical study. It could serve as an example for a preregistration (or any other type of study planning or proposal).

You plan to measure 100 features in a population that you think constitutes two equally sized subgroups. Between these subgroups, you expect small differences (Cohen's $d$ = 0.3) in 20 features, medium differences (Cohen's $d$ = 0.5) in 12, and large differences (Cohen's $d$ = 0.8) in 4. With these differences among 36/100 measured features, total expected separation would be Δ=2.7 (computed with Equation 1).

Using multi-dimensional scaling, the expected separation of Δ=2.7 in original space would be expected to increase to roughly Δ=4 in reduced (two-dimensional) space (according to Figure 7). This increased separation offers better power for a following cluster analysis.

To further improve your statistical power, you can choose to work with a fuzzy clustering algorithm (e.g. c-means) or a mixture modelling approach (e.g. latent profile or latent class analysis, depending on whether your data is continuous or binary), which both showed sufficient power at the expected separation of Δ=4 (according to Table 4, and Figures 10, 12, and 13).

The expected separation and choice of analysis pipeline provides acceptable statistical power if you measure 20 to 30 observations per expected subgroup (according to Table 4 and Figure 10). You expected two equally sized subgroups, so your total sample size would be N=40 to N=60.

If you had expected one smaller and one larger subgroup, for example 20% and 80% of the population, this would need to be accounted for in your sample size. Your sample should include at least 20 to 30 observations from the smaller subgroup. Assuming unbiased sampling (20% of your sample is part of the smaller subgroup in the population), your total sample should thus be N=100 to N=150. You can compute this as: (100% / subgroup percentage) * minimum number of samples.

## Conclusion

Cluster algorithms have sufficient statistical power to detect subgroups (multivariate normal distributions with different centres) only when these are sufficiently separated. Specifically, the separation in standardised space (here named effect size $\Delta$) should be at least 4 for k-means and HDBSCAN to achieve over 80% power. Better power is observed for c-means and finite Gaussian mixture modelling, which achieved 80% power at $\Delta=3$ without inflating the false positive rate. Effect size $\Delta$ can be computed as the accumulation of expected within-feature effect sizes. While covariance structure did not impact clustering, sample size did to some extent, particularly for HDBSCAN and k-means. Sampling at least N=20 to 30 observations per expected subgroup resulted in satisfactory statistical power.

# Methods

## Literature frequency comparison

Figure 1 in the Background section describes an increase in the relative frequency of publications that mention several terms related to cluster analysis, compared to publications that mention traditional statistics. The illustration was generated using Bibliobanana (17), which uses the NCBI E-utilities API to query the publications that appeared in the PubMed database each year.

Search queries used the TEXT field (keyword could appear anywhere in a publication), and were "k-means" (k-means, blue line in Figure 1), "agglomerative clustering" OR "hierarchical clustering" (hierarchical clustering, green line), "DBSCAN" OR "HDBSCAN" ((H)DBSCAN, purple line), "c-means" (c-means, red line), "finite mixture modelling" OR "latent class analysis" OR "latent profile analysis" (finite mixture modelling, orange line), and "t-SNE" OR "tSNE" OR "UMAP" (t-SNE / UMAP, brown line). Comparison terms were "t-test", "ANOVA", and "linear regression". Clustering terms were divided by the average of all comparison terms to come to their relative frequency.

## Simulation

All simulated datasets constituted multivariate normal distributions, specifically one for each subgroup. We could define the covariance structure of each distribution, and distribution separation by setting their mean vectors. All standard deviations were set to 1.

For the first simulation analysis, generated datasets constituted 1000 observations and 15 features, with defined mean vectors (see below), and standard deviations of 1 within each feature. The number of subgroups was 2 with unequal group size (10/90%), 2 with equal group size (50/50%), or 3 with equal group size (33/34/33%). Within each simulation, within-feature differences were generated with Cohen's *d* values of 0.3, 0.5, 0.8, 1.3, or 2.1, and the number of different features was 1, 5, 10, or 15.

For two-cluster datasets, centroids (mean vectors) were determined by subtracting half the intended Cohen's *d* from one cluster, and adding the same value to the other within each feature. The order of addition and subtraction was shuffled within each feature. For three-cluster datasets, one cluster was assigned a mean vector of zeros (the "middle" cluster). The other two clusters had the intended Cohen's *d* added or subtracted within each feature, again with shuffled order within each feature.

Covariance structures were generated in three different ways: no covariance with values of 0 for all non-diagonal relations; uniform random values between -0.3 and 0.3; or with imposed 3 or 4 factor structure with uniform random values between -0.9 and -0.4 or between 0.4 and 0.9 within each factor, and between -0.3 and 0.3 for relations between variables in different factors. We also included two types of datasets with different covariance structures between subgroups, one with different underlying factor structures (3 and 4 factors; or 3 and 4 and no factors), and one simply with different covariance structures (random and random; or random and random and no covariance).

To compute statistical power and accuracy, further simulations were run to generate datasets of 10, 20, 40, 80, or 160 observations and 2 (uncorrelated) features. These datasets were constructed with two unequally sized (10/90%), two equally sized, three equally sized, or four equally sized subgroups (multivariate normal distributions with standard deviations of one); all with equidistant means at centroid separations of $\Delta=1$ to 10. For each combination of variables, new data were generated in 100 iterations, and then analysed with k-means, HDBSCAN, or c-means (see below).

Finally, to compare the power and accuracy of k-means, c-means, and Gaussian mixture modelling, we simulated datasets of 120 observations and 2 (uncorrelated) features with one, two, three, or four equally sized and equidistant multivariate normal distributions. Their separations varied from $\Delta=1$ to 10, and they were analysed with k-means, c-means, and Gaussian mixture modelling (see below).

## Open Code and Data

Data was simulated in Python (37) (version 2.7.12; for a tutorial, see reference (38)), using the NumPy package (version 1.16.5) (39,40). Dimensionality reduction and clustering was performed using the packages SciPy (version 1.2.2) (39), umap-learn (version 0.2.1) (31,41), hdbscan (version 0.8.12) (34), scikit-fuzzy (version 0.4.2), and scikit-learn (version 0.20.4) (2). Outcomes were plotted using Matplotlib (version 2.1.2) (42). All code and simulated data used for this manuscript can be found on GitHub, from where it can be freely accessed and downloaded: www.github.com/esdalmaijer/cluster_power

The linked resource also contains additional plots for specific simulations not illustrated here, and can be altered to test additional types of dimensionality reduction and clustering algorithms. We have already implemented 13 further dimensionality reduction algorithms, and 8 additional cluster algorithms. Researchers with a special interest in any of these are welcome to use our resource to compute statistical power for their specific situation.

## **Dimensionality reduction**

Cluster analysis is usually performed on high-dimensional data, i.e. with many measured features per observation. While it is possible to apply clustering algorithms directly, the "curse of dimensionality" entails that this approach is unlikely to yield strong results (26). Instead, many opt for projecting high-dimensional data into a lower-dimensional space. One option for this is principal component analysis (PCA), but extracting only a few components risks removing meaningful variance. Instead, data can be projected in two-dimensional space with limited loss of information with multi-dimensional scaling (30) (MDS), a technique that aims to retain inter-observation distances in original data in a lower-dimensional projection. Finally, algorithms such as t-stochastic neighbour embedding (t-SNE) (43) and uniform manifold approximation and projection (UMAP) (31) non-linearly reduce dimensionality, effectively retaining local inter-sample distances while exaggerating global distances. An additional advantage of these techniques is that data projected into two or three dimensions can be plotted, and thus visually inspected for oddities, and perhaps even provide a rough indication of grouping.

We employed three reduction strategies: None, MDS, and UMAP.

## Clustering

After dimensionality reduction, the resulting dataset can be subjected to a wide selection of clustering algorithms that each have optimal conditions (for an overview, see (44)). Here, we will explore the most common types. This includes k-means (32), an algorithm that arbitrarily draws a predefined number (k) of centroids within the data, and on each iteration moves the centroids to the average of the observations that are closest to each centroid, until a stable solution is reached. Another approach is agglomerative (hierarchical) clustering, which recursively joins pairs of observations according to a combination of linkage affinity (e.g. Euclidean or cosine distance) and criterion. A commonly used linkage is Ward, which minimises the variance of merging groups of observations (33). Because these algorithms require the user to define the number of clusters, a common approach is to cycle through a variety of options to identify the best fitting solution.

A class of algorithms that does not require the prespecification of an expected number of clusters includes DBSCAN (45) and HDBSCAN (34). They identify clusters of denser observations among lower-density observations that remain unassigned.

We employed five algorithms: k-means, agglomerative clustering with Ward linkage and Euclidean distance, agglomerative clustering with average linkage and cosine distance, HDBSCAN, and c-means (fuzzy clustering; see below).

## Outcome evaluation

After observations are assigned a cluster, the quality of the solution can be determined. For each sample, a silhouette coefficient can be computed as the relative distance to its assigned centroid and the nearest other centroid (46). For each observation, a value of 1 means perfect alignment with its assigned centroid, 0 means it lies exactly in between its centroid and the nearest other, and -1 means perfect alignment with a centroid it was not assigned to. The average across all assigned observations is the silhouette score, which is often taken as evidence for clustering if it exceeds 0.5, or as strong evidence if it exceeds 0.7 (47). It should be noted that there are many cluster validation indices (for excellent overviews, see (19,48)). We focus on the silhouette score because of its good performance in many circumstances (19), conceptual elegance, and established thresholds for interpretation. Unassigned observations (such as in HDBSCAN) are ignored for silhouette score computation. Scores were computed slightly differently for fuzzy clustering tools (see section "*Fuzzy Clustering*" below), but interpreted in the same framework.

While a ground truth is normally not available, it is in the context of simulated data. This allowed us to compute the Rand index (49), adjusted for chance (50), to quantify the overlap between cluster outcome and ground truth. An adjusted Rand index of 1 reflects perfect match, a value of 0 means chance performance, and negative values indicate the clustering performed worse than chance. While the adjusted Rand index quantifies the overlap between cluster outcome and truth, the silhouette coefficient reflects what an experimenter (who is normally blind to the ground truth) would conclude.

**Table 5**

*Outcome evaluation metrics and their meanings.*

| Value | Silhouette score meaning (Computed from data) | Adjusted Rand index meaning (Computed from data and ground truth) |
|---|---|---|
| < 0 | Observations are closer to the centroid of another cluster than to the centroid of the cluster to which they were assigned. | Below chance performance: random guessing would have provided more accurate labelling. |
| 0 | Observations lie exactly between the centroids of their assigned cluster and the nearest of the other clusters. | The number of labels assigned by the algorithm that overlap with the ground truth are at chance level. |
| 0.5 – 0.7 | Evidence for clustering | |
| 0.7 – 1.0 | Strong evidence for clustering | |
| 1 | Perfect alignment of observations and their assigned cluster centroid. | Perfect overlap between assigned clusters and ground truth. |

# Fuzzy clustering and mixture modelling

We employed the c-means algorithm (51,52), specifically the version implemented in Python package scikit-fuzzy (53). It converges on centroids in a similar way to k-means, but allows for observations to be assigned to more than one cluster. Specifically, each observation is assigned $k$ values between 0 and 1 that indicate membership likelihood.

In addition to c-means, we employed finite mixture modelling through scikit-learn's GaussianMixture class (2). This approach aims to find the best mixture of $k$ Gaussian distributions, allowing each component their own general covariance matrix. The resulting model was used to compute for each observation the probability that it was part of each Gaussian; a similar outcome to the c-means algorithm.

We estimated c-means and mixture model outcomes using a variation of the silhouette coefficient intended for fuzzy clustering methods (54). This silhouette score has a tunable exponentiation parameter α that determines how strongly the uncertainty about each observations cluster membership is weighted (when it approaches 0, the fuzzy silhouette coefficient approaches the regular version), which was set to 1 in our analyses.

For mixture modelling, we also computed the Bayesian Information Criterion (BIC) for each fitted solution. Where the silhouette score should be maximised to identify the best cluster solution, the BIC should be minimised.

As described above, datasets (N=120) were simulated with 1-4 subgroups (multivariate normal distributions with SD=1) with separations of Δ=1 to 10. A new dataset was simulated in 100 iterations. In each, k-means and c-means were applied, with predefined guesses of k=2 to 7. From the outcomes, we computed the probability of each analysis to detect clustering (silhouette coefficient >= 0.5), and to detect the correct number of clusters (silhouette coefficient highest for the value of k that corresponded with ground truth).

## Effect of dimensionality reduction on cluster separation

In the simulated datasets, distances between cluster centroids in original feature space should be Euclidean (Equation 1). However, due to the stochastic and non-linear nature of dimensionality reduction algorithms, centroid distances after dimensionality reduction are less predictable. To quantify the effect of dimensionality reduction on cluster separation, we computed the distance between cluster centroids (defined as the average Euclidean position of observations within a cluster) in projected space after dimensionality reduction.

(1) $$\Delta = \sqrt{\sum_{i=1}^{n} \delta_i^2}$$

Where Δ is centroid distance, $n$ is the number of features, and δ is the within-feature difference between clusters (effect size).

# Introducing effect size Δ

We consider multivariate normal distributions with standard deviations of 1 (for all features) to be standardised space, and refer to cluster separation in this space as Δ. It can serve as an effect size metric for clustering in the sense that it reflects the extent of separation of simulated or identified subgroups. It is essentially the multivariate equivalent of Cohen's d, and can in fact be estimated from expected values of Cohen's d within each feature via Equation 1.

# Power and accuracy

Researchers who opt for cluster analysis are likely attempting to answer three main questions: 1) Are subgroups present in my data, 2) How many subgroups are present in my data, and 3) Which observations belong to what subgroup?

In null-hypothesis testing, power relates to the probability that a null hypothesis is correctly rejected if an alternative hypothesis is true. Various approaches have been suggested to define statistical power in cluster analyses, for example through outcome permutation (55) or through measures of subgroup overlap (56). Here, we define power as the likelihood of a cluster analysis to accurately reject the null hypothesis that no subgroups are present, based on the binary decision of a solution's silhouette score being 0.5 or over (47).

Further to the binary decision of clusters being present in a dataset, we estimated the probability of cluster analyses to identify the correct number of subgroups in simulated datasets. This was done by cycling through k=2 to k=5 for algorithms that require pre-specification of cluster number (k-means and agglomerative clustering), and choosing the value that resulted in the highest silhouette coefficient. HDBSCAN reports the number of detected clusters, and thus did not require iterating through pre-specified values.

Finally, we quantified classification accuracy as the proportion of observations correctly assigned to their respective clusters. This reflects the overlap between ground truth and assigned cluster membership. This accuracy is dependent on chance, which was set at the proportion of the total sample size that was in the largest cluster.

# Declarations

### Ethics approval and consent to participate

No participants were tested for this study, which relies on simulations only.

### Availability of data and materials

The authors have made their data and analysis software available publicly. It can be accessed on GitHub: [https://github.com/esdalmaijer/cluster_power/](https://github.com/esdalmaijer/cluster_power/)

### Competing interests

The authors declare that they have no competing interests.

### Funding

ESD and DEA are supported by grant TWCF0159 from the Templeton World Charity Foundation to DEA. CLN is supported by an AXA Fellowship. All authors are supported by the UK Medical Research Council, grant MC-A0606-5PQ41.

## Authors' contributions

ESD and CLN first initiated the study. ESD conceptualised the study, developed the methods, simulated and analysed the data, and drafted the manuscript. All authors interpreted the results, provided critical feedback on the manuscript, and approved of the final version.